# Discriminative Supervised Hashing for Cross-Modal similarity Search

Jun Yu[a], Xiao-Jun Wu[a, *] and Josef Kittler[b]

[a]*School of Internet of Things Engineering, Jiangnan University, Wuxi 214122, China*
[b]*CVSSP, University of Surrey, Guildford, GU2 7XH, UK*

## ABSTRACT

With the advantage of low storage cost and high retrieval efficiency, hashing techniques have recently been an emerging topic in cross-modal similarity search. As multiple modal data reflect similar semantic content, many researches aim at learning unified binary codes. However, discriminative hashing features learned by these methods are not adequate. This results in lower accuracy and robustness. We propose a novel hashing learning framework which jointly performs classifier learning, subspace learning and matrix factorization to preserve class-specific semantic content, termed Discriminative Supervised Hashing (DSH), to learn the discriminative unified binary codes for multi-modal data. Besides, reducing the loss of information and preserving the non-linear structure of data, DSH non-linearly projects different modalities into the common space in which the similarity among heterogeneous data points can be measured. Extensive experiments conducted on the three publicly available datasets demonstrate that the framework proposed in this paper outperforms several state-of-the-art methods.

## 1. Introduction

Recently, the explosion of multimedia data, such as image, text, video, and audio, increases the demands for high efficiency, low storage cost and effectiveness of retrieval applications. Hashing has received much attention in information retrieval and related areas because of the retrieval processing speed. Among many hashing methods [1-8], Minimal loss hashing (MLH) [2] is a framework based on the latent structural SVM. Kernel-based supervised hashing (KSH) [3], Supervised discrete hashing with point-wise labels (SDH) [1] and Scalable discrete hashing with pairwise supervision (COSDISH) [7] have been shown to deliver reasonable retrieval performance. However, the above methods have been designed for unimodal data setting and are not directly applicable to cross-modal retrieval.

Cross-modal is a very interesting scenario. For example, for a given image, it may be possible to retrieve semantically relevant texts from the database. But, it is hard to directly measure the similarity between different modalities. To tackle the problem, most existing methods [9-14] focus on finding a common subspace where the heterogeneous data can be measured. For instance, the main idea of the Inter-Media Hashing (IMH) [10] is that two points from the same neighborhood should be as close as possible in the common subspace. Semi-Paired Discrete Hashing (SPDH) [13] explore the common latent subspace by constructing a cross-view similarity graph. Fusion Similarity Hashing (FSH) [9] learns the hashing function by preserving the fusion similarity. However, the learned hashing codes have weak discrimination ability.

Benefitting from the discriminative information provided by category labels, supervised hashing methods [15-18] often improve the retrieval accuracy. Cross View Hashing (CVH) [17] aims to minimize the hamming distance between data objects belonging to the same class in a common hamming space. Semantic Correlation Maximization (SCM) [15] learns discriminative binary codes based on the cosine similarity between the semantic label vectors. Supervised Matrix Factorization Hashing (SMFH) [18] integrates graph regularization into the hashing learning framework. However, they tend to learn hashing through preserving the similarities of the inter-modal and intra-modal data but cannot ensure the learned hashing codes are semantically discriminative. In fact, it is very important that those samples with the same label have similar binary codes for cross-modal similarity search. Moreover, the computational cost of similarities of the inter-modal and intra-modal data is relatively high.

To tackle the problem, we propose a DSH model which integrates the classifier learning and matrix factorization with consistent label into hashing learning framework. Furthermore, kernelized hash functions are learned for out-of-sample extension. Fig. 1 illustrates the overall framework of the proposed DSH. Compared with [19], our framework explores the shared structure of each category. The main contributions of DSH hashing method are given as follows:

1) To learn more discriminative binary codes, DSH learns unified binary codes by combining classifier learning and label consistent matrix factorization.

---

* Corresponding author. Tel.: +86-13601487486; e-mail: wu_xiaojun@jiangnan.edu.cn

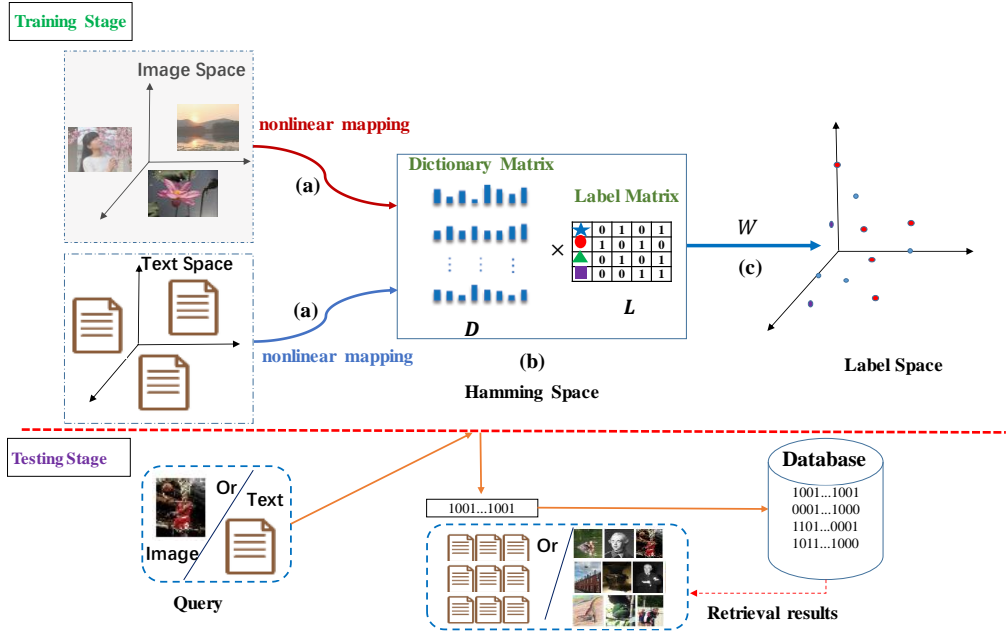

**Fig. 1.** The illustration of DSH. (a) Image and Text from isomeric space are mapped nonlinearly into the common Hamming space respectively; (b) In the Hamming space, the common features preserve class-specific semantic information. (c) The category information of the samples represented by the common hashing features are predicated. In testing phase, we can obtain the hashing codes of an arbitrary query (an image or a text), and semantically related another modal data (texts or images) are returned from the Database.

2) DSH learns hashing functions for each modality through employing the kernel method which can capture non-linear structural information of object.

Structurally, the rest of this paper falls into three sections. Our model and optimization algorithm are presented in the Section 2. Section 3 shows the experimental results on three available datasets. Finally, the conclusions are drawn in the section 4. The source code of DSH proposed in this paper is available.

## 2. Our Methodology

### 2.1. Notation and Problem Statement

Suppose that $O = [o_1, o_2, ..., o_n]$ is a set of $n$ training instances with $m$ modalities pairs. $X^{(m)} = [x_1^{(m)}, x_2^{(m)}, ..., x_n^{(m)}]$ denotes the $m$-th modality, where $x_i^{(m)} \in R^{d_m}$ is the $i$-th sample of $X^{(m)}$ with dimension $d_m$. $L = \{l_1, l_2, ..., l_n\} \in R^{C \times n}$ is a label matrix, where $C$ denotes the number of categories. $l_{ik}$ is the $k$-th element of $l_i$, $l_{ik} = 1$ if the $i$-th instance belongs to the $k$-th category and $l_{ik} = 0$ otherwise. Here an instance $o_1$ can be classified into multiple categories. Without loss of generality, data points are zero-centered for each modality, i.e. $\sum_i^n X^{(m)} = 0$. The aim of the DSH proposed in this paper is to learn a mapping function $f^m(\cdot) = \{f_1^m(\cdot), f_2^m(\cdot), ..., f_r^m(\cdot)\}$, where $r$ is the length of binary codes, from the original data space to Hamming space for the $m$-th modality. Taking the $i$-th sample of the $m$-th modality for example, the $j$-th element of hash code is obtained as follows

$$h_j(x_i^{(m)}) = sgn(f_j^m(x_i^{(m)})), j=1,2,...,r \quad (1)$$

where $sgn(.)$ is the sign function which is equivalent to '1' when its argument is positive and '0' otherwise.

### 2.2. Common space learning

It is difficult to directly measure the similarity of two data points (such as an image and a sentence) from different feature spaces. They should be relatively close if they share similar semantic information content. We are interested in learning a latent common subspace where the similarity between different modalities can be calculated using the Euclidean distance. However a linear embedding cannot preserve manifold structure among data points. Inspired by [3, 20, 21], we adopt nonlinear mapping to project isomeric modalities into the common space.

$$H^m(x_i^{(m)}) = P^{(m)} \kappa(x_i^{(m)}) \quad (2)$$

Here we employ RBF kernel function to calculate the kernel matrix $\kappa(x) = [\exp(-||x-a_1||^2/\sigma), ..., \exp(-||x-a_M||^2/\sigma)]^T$, where $\{a_i\}_{i=1}^M$ denotes kernel-bases with sampling size $M$ and $\sigma$ is the kernel width. $P^{(m)} \in R^{r \times M}$ maps the generated $\kappa(x_i^{(m)})$ into latent common space.

### 2.3. Discrimination Preservation

We hope that the learned hashing codes of samples classified into different categories can be clearly distinguished. In other words, the hashing feature are discriminative enough. Accordingly, the category of all samples can be predicted effectively through minimizing the loss of classification. If an image-text pair reflects consistent information semantically, they should share the same binary code. The objective function of learning discriminative unified binary codes is

$$\min_B ||WB - L||^2 + \lambda ||W||^2 \quad (3)$$

where $L \in R^{c \times n}$ is label matrix, $c$ is the number of categories. $W \in R^{c \times r}$ is a linear multi-class classifier and $\lambda$ is a balance parameter.

### 2.4. Class-Specific semantic Preserving

If a sample belongs to a specific category, we assume it has overall attributes of the category. That is to say, all samples classified into specific category share common features. Thus, for a data point classified into multiple classes, it is natural that the data point combines the common features of multiple categories. To further utilize the label information and better represent data, DSH learns a discriminative basis matrix using the semantic label information. Specifically, we reconstruct the unified representation in hamming space, i.e. $B \approx DL$, where $D \in R^{r \times c}$

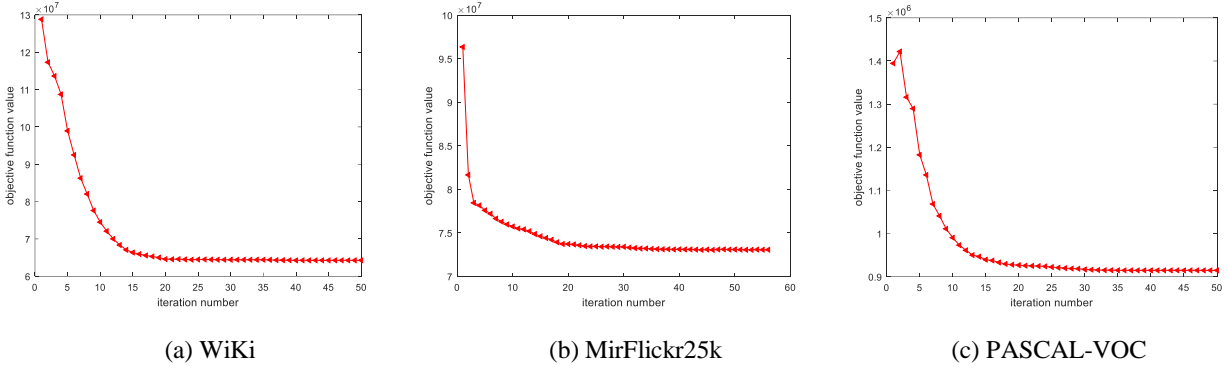

**Fig. 2**. The convergence of algorithm 1 on WiKi (a), MirFlickr25k (b) and PASCAL-VOC (c).

is a basis matrix with $c$ columns. The reconstruction error is defined as

$$\min_D ||B - DL||^2 \qquad (4)$$

We integrate the common space learning in (2), discrimination preserving objective in (3) and category-specific semantics preserving goal in (4) into a joint optimization problem. The overall objective function of DSH is defined as

$$\min_{B,D,W,P^{(m)}} \sum_{m=1}^{v} (\alpha^{(m)})^\gamma \{||B - sgn(P^{(m)}\kappa(X^{(m)}))||_F^2\} \\ + \beta||WB - L||^2 + \eta||B - DL||^2 \\ + \lambda\varphi(D, W, P^{(m)}) \\ s.t. \sum_{v=1}^{m} \alpha^{(m)} = 1, B \in \{-1,1\}^{r \times n} \qquad (5)$$

where $\varphi(D, W, P^{(m)}) = ||D||^2 + ||W||^2 + \sum_{v=1}^{m}||P^{(v)}||^2$ is the regularization term to avoid overfitting, $\alpha^{(m)}$ is a weight factor of the $m$-th modality and the role of $\gamma$ is to smoothen the weight distribution, $\beta$ and $\lambda$ are non-negative balance parameters. Rather than learning $P^{(1)}, P^{(2)} ... P^{(v)}$ separately, we learn these mapping functions simultaneously to get the optimal solution globally.

*2.5. Optimization*

The problem in (5) is not differentiable because of the sign function. We relax the objective function by ignoring the sign function. Then the relaxed problem can be rewritten as

$$\min_{B,D,W,P^{(m)}} \sum_{m=1}^{v} (\alpha^{(m)})^\gamma \{||B - P^{(m)}\kappa(X^{(m)})||_F^2\} \\ + \beta||WB - L||^2 + \eta||B - DL||^2 \\ + \lambda\varphi(D, W, P^{(m)}) \\ s.t. \sum_{v=1}^{m} \alpha^{(m)} = 1, B \in \{-1,1\}^{r \times n} \qquad (6)$$

The optimization problem in (6) with multiple variables is not convex. We introduce ADMM algorithm to update every variable while keeping the other variables fixed. The detailed optimization procedure is presented as follows

**Update W with other variables fixed.** The objective function (6) can be rewritten as the follows

$$\min_W \beta||WB - L||^2 + \lambda||W||^2 \qquad (7)$$

Clearly, (7) has the closed solution

$$W = \beta LB^T(\beta BB^T + \lambda I)^{-1} \qquad (8)$$

**Update $P^{(m)}$ with other variables fixed.** Keeping only the terms relating to $P^{(m)}$, we can obtain

$$\min_{P^{(m)}} (\alpha^{(m)})^\gamma \{||B - P^{(m)}\kappa(X^{(m)})||_F^2\} + \lambda||P^{(m)}||^2 \qquad (9)$$

By setting the derivative of (9) w.r.t. $P^{(m)}$ to zero, $P^{(m)}$ can be updated as follows

**Algorithm 1:** Discriminative Supervised Hashing (DSH)

**Input**: training set $X^{(m)} \in R^{d_x \times n}$, label matrix $L \in \{0,1\}^{C \times n}$, and parameters $\beta, \lambda$.
**Output**: projection matrix $P^{(m)} \in R^{r \times M}$
1: Initialize $B, P^{(m)}, Z, W, \alpha^{(m)}, \gamma$.
2: Calculate $\kappa(X^{(m)})$ for all modalities
4: **Repeat**
5:   Update $W$ according to (8)
6:   Update $P^{(m)}$ according to (10);
7:   Update $D$ according to (12);
8:   Update $B$ according to (15);
9:   Update $\alpha^{(m)}$ by solving (18);
10: **Until** convergence
11: Return $B, P^{(m)}, Z, W, \alpha^{(m)}$.

$$P^{(m)} = (\alpha^{(m)})^\gamma B\kappa(X^{(m)^T}((\alpha^{(m)})^\gamma \kappa(X^{(m)}\kappa(X^{(m)^T} + \lambda I)^{-1} \qquad (10)$$

**Update $D$ with other variables fixed.** The sub-problem is to minimize the following function:

$$\min_D \eta||B - DL||^2 + \lambda||D||^2 \qquad (11)$$

Similar to $W$, $D$ is updated as follows

$$D = \eta BL^T(\eta LL^T + \lambda I)^{-1} \qquad (12)$$

**Update $B$ with other variables fixed.** The sub-problem w.r.t. $B$ is written as

$$\min_B \sum_{m=1}^{v} (\alpha^{(m)})^\gamma ||B - P^{(m)}\kappa(X^{(m)})||_F^2 + \beta||WB - L||^2 \\ + \eta||B - DL||^2 \\ s.t. \quad B \in \{-1,1\}^{r \times n} \qquad (13)$$

The above problem with discrete constraints is NP hard. One common method to solve it is to apply relaxation strategy. We introduce the discrete cyclic coordinate descent (DCC) algorithm [1] to learn a specific row $b^T$ of binary codes $B$ by fixing the other rows. Then (13) can be written as

$$\min_B -2tr(QB) + \beta||WB||^2 \qquad (14)$$

where $Q = \sum_{m=1}^{v} (\alpha^{(m)})^\gamma \kappa(X^{(m)^T}P^{(m)^T} + \beta L^T W + \eta L^T D^T$. Then $b$ has the closed form solution

$$b = sgn(q - \bar{B}^T \bar{W}^T w) \qquad (15)$$

Where $q$ and $w$ denote a column of $W$ and $Q$ respectively, $\bar{W}$ and $\bar{B}$ are two matrices excluding $w$ and $b^T$ respectively.

**Update $\alpha^{(m)}$ with other variables fixed.** The sub-problem of $\alpha^{(m)}$ is:

$$\min_{\alpha^{(v)}} \sum_{m=1}^{v} (\alpha^{(m)})^{\gamma} C^{(m)}$$
$$s.t. \sum_{m=1}^{v} \alpha^{(m)} = 1 \quad (16)$$

where $C^{(m)} = ||B - P^{(m)} \kappa(X^{(m)})||_F^2$. The problem is transformed into (17) by the method of Lagrange multipliers

$$\min_{\alpha^{(m)}} \sum_{m=1}^{v} (\alpha^{(m)})^{\gamma} C^{(m)} + \xi \left(1 - \sum_{m=1}^{v} \alpha^{(m)}\right) \quad (17)$$

Taking the derivative of (17) with respect to $\alpha^{(m)}$ set to zero, we get

$$\alpha^{(m)} = \frac{(\gamma C^{(m)})^{1/(1-\gamma)}}{\sum_{m=1}^{v} (\gamma C^{(m)})^{1/(1-\gamma)}} \quad (18)$$

The detailed optimization procedure is summarized in Algorithm 1. The process is repeated until the algorithm converges.

*2.6. Generating Hash Codes*

For an unseen sample, we can generate a binary code by the learned hash function. Given a query $x$ from the m-th modality, we first calculate the kernel vector $\kappa(x)$, and then the binary code $b$ is generated by the mapping $b = sgn(P^{(m)} \kappa(x))$.

**Table 1 MAP results of I2A and A2I on WiKi**

| Task | Method | Code Length | | | |
|------|--------|-------|-------|-------|-------|
|      |        | 16 | 32 | 64 | 128 |
| I2T | CCA | 0.1699 | 0.1519 | 0.1495 | 0.1472 |
|     | IMH | 0.2022 | 0.2127 | 0.2164 | 0.2191 |
|     | SCM-orth | 0.1538 | 0.1402 | 0.1303 | 0.1289 |
|     | SCM-seq | 0.2341 | 0.2410 | 0.2462 | 0.2566 |
|     | SMFH | 0.1763 | 0.2409 | 0.2539 | 0.2564 |
|     | FSH | 0.2346 | 0.2491 | 0.2549 | 0.2573 |
|     | DCH-RBF | 0.2385 | 0.2495 | 0.2725 | 0.2759 |
|     | DSH$_{linear}$ | 0.2398 | 0.2631 | 0.2721 | 0.2693 |
|     | DSH | 0.2593 | 0.2748 | 0.2853 | 0.2929 |
| T2I | CCA | 0.1587 | 0.1392 | 0.1272 | 0.1211 |
|     | IMH | 0.1648 | 0.1703 | 0.1737 | 0.1720 |
|     | SCM-orth | 0.1540 | 0.1373 | 0.1258 | 0.1224 |
|     | SCM-seq | 0.2257 | 0.2459 | 0.2485 | 0.2528 |
|     | SMFH | 0.3821 | 0.5716 | 0.5868 | 0.6053 |
|     | FSH | 0.2149 | 0.2241 | 0.2298 | 0.2368 |
|     | DCH-RBF | 0.6984 | 0.7162 | 0.7223 | 0.7229 |
|     | DSH$_{linear}$ | 0.2463 | 0.2637 | 0.2753 | 0.2791 |
|     | DSH | 0.7266 | 0.7486 | 0.7553 | 0.7636 |

**Table 2 MAP results of I2T and T2I on MirFlickr 25k**

| Task | Method | Code Length | | | |
|------|--------|-------|-------|-------|-------|
|      |        | 16 | 32 | 64 | 128 |
| I2T | CCA | 0.5744 | 0.5706 | 0.5681 | 0.5658 |
|     | IMH | 0.5821 | 0.5825 | 0.5810 | 0.5774 |
|     | SCM-orth | 0.5884 | 0.5743 | 0.5678 | 0.5659 |
|     | SCM-seq | 0.6222 | 0.6292 | 0.6391 | 0.6440 |
|     | SMFH | 0.6101 | 0.6238 | 0.6128 | 0.6122 |
|     | FSH | 0.6118 | 0.6139 | 0.6263 | 0.6268 |
|     | DCH-RBF | 0.6193 | 0.6194 | 0.6145 | 0.6215 |
|     | DSH$_{linear}$ | 0.6242 | 0.6325 | 0.6437 | 0.6525 |
|     | DSH | 0.6257 | 0.6396 | 0.6462 | 0.6537 |
| T2I | CCA | 0.5737 | 0.5701 | 0.5678 | 0.5660 |
|     | IMH | 0.5799 | 0.5810 | 0.5816 | 0.5788 |
|     | SCM-orth | 0.5848 | 0.5743 | 0.5686 | 0.5652 |
|     | SCM-seq | 0.6128 | 0.6182 | 0.6241 | 0.6304 |
|     | SMFH | 0.6105 | 0.6224 | 0.6117 | 0.6087 |
|     | FSH | 0.6035 | 0.6057 | 0.6133 | 0.6137 |
|     | DCH-RBF | 0.6065 | 0.6104 | 0.6091 | 0.6158 |
|     | DSH$_{linear}$ | 0.6136 | 0.6233 | 0.6385 | 0.6418 |
|     | DSH | 0.6267 | 0.6404 | 0.6545 | 0.6603 |

**Table 3 MAP results of I2T and T2I on PASCAL-VOC**

| Task | Method | Code Length | | | |
|------|--------|-------|-------|-------|-------|
|      |        | 16 | 32 | 64 | 128 |
| I2T | CCA | 0.1245 | 0.1267 | 0.1230 | 0.1218 |
|     | IMH | 0.2087 | 0.2016 | 0.1873 | 0.1718 |
|     | SCM-orth | 0.1565 | 0.1383 | 0.1282 | 0.1214 |
|     | SCM-seq | 0.2554 | 0.3253 | 0.2451 | 0.3388 |
|     | SMFH | 0.1924 | 0.2213 | 0.2600 | 0.2844 |
|     | FSH | 0.2831 | 0.3237 | 0.3340 | 0.3496 |
|     | DCH-RBF | 0.2590 | 0.3572 | 0.3633 | 0.3776 |
|     | DSH$_{linear}$ | 0.2791 | 0.3203 | 0.3517 | 0.3555 |
|     | DSH | 0.4113 | 0.4217 | 0.4476 | 0.4566 |
| T2I | CCA | 0.1283 | 0.1362 | 0.1465 | 0.1553 |
|     | IMH | 0.1631 | 0.1558 | 0.1537 | 0.1464 |
|     | SCM-orth | 0.1982 | 0.1501 | 0.1211 | 0.1018 |
|     | SCM-seq | 0.2989 | 0.4108 | 0.2652 | 0.4531 |
|     | SMFH | 0.3874 | 0.5081 | 0.5986 | 0.6609 |
|     | FSH | 0.2574 | 0.3030 | 0.3216 | 0.3428 |
|     | DCH-RBF | 0.3278 | 0.5458 | 0.6055 | 0.6253 |
|     | DSH$_{linear}$ | 0.3950 | 0.4922 | 0.5824 | 0.6209 |
|     | DSH | 0.8620 | 0.8814 | 0.9003 | 0.9087 |

**Table 4 The statistics of different sampling size**

| Dataset | | Sampling Size | | | | |
|---------|---|------|------|------|------|------|
|         |   | 100 | 600 | 1100 | 1600 | 2100 |
| WiKi | I2T | 0.2289 | 0.2732 | 0.2592 | 0.2439 | 0.2593 |
|      | T2I | 0.2214 | 0.4790 | 0.6117 | 0.6832 | 0.7266 |
|      | time (s) | 0.1461 | 1.5184 | 4.1867 | 5.1870 | 21.8003 |
| MirFlickr25k | I2T | 0.5933 | 0.6217 | 0.6156 | 0.6127 | 0.6257 |
|      | T2I | 0.5959 | 0.6189 | 0.6199 | 0.6230 | 0.6267 |
|      | time (s) | 1.4815 | 7.3464 | 18.9943 | 35.4156 | 47.9027 |
| PASCAL-VOC | I2T | 0.3563 | 0.3895 | 0.4128 | 0.4039 | 0.4113 |
|      | T2I | 0.2752 | 0.5844 | 0.6966 | 0.7906 | 0.8620 |
|      | time (s) | 0.0582 | 0.6264 | 1.6915 | 4.1216 | 7.3768 |

## 3. Experiments and analysis

*3.1. Datasets*

**Wiki** [22] contains 2,866 multimedia documents harvested from Wikipedia. Every document consists of a pair of an image and a text description, and every paired sample is classified as one of 10 categories. We take 2,866 pairs from the dataset to form the training set and the rest as a test set.

**MirFlickr25k** dataset [23] is collected from Flickr website. It consists of 25,000 image-text pairs and each pair is assigned into some of 20 categories. We keep 20,015 pairs which have at least 20 textual tags for our experiments. Each image is represented by 150-dimensional edge histogram features, and each text is represented by 500-dimensional feature vector which is extracted from Bag-of-word representation employing PCA. We randomly select 5,000 pairs as a training set from a 15,902 retrieval set and the rest as a query set.

**PASCAL-VOC** [24] consists of 9,963 image-tag pairs. Each image is represented by a 512-dimensional Gist Feature vector and each text is represented as 399-dimensional word frequency count. All pairs are classified into 20 different categories. In our experiment, we select 5,649 images with only one object. 2,808 pairs are taken out as a training set and the remaining as query.

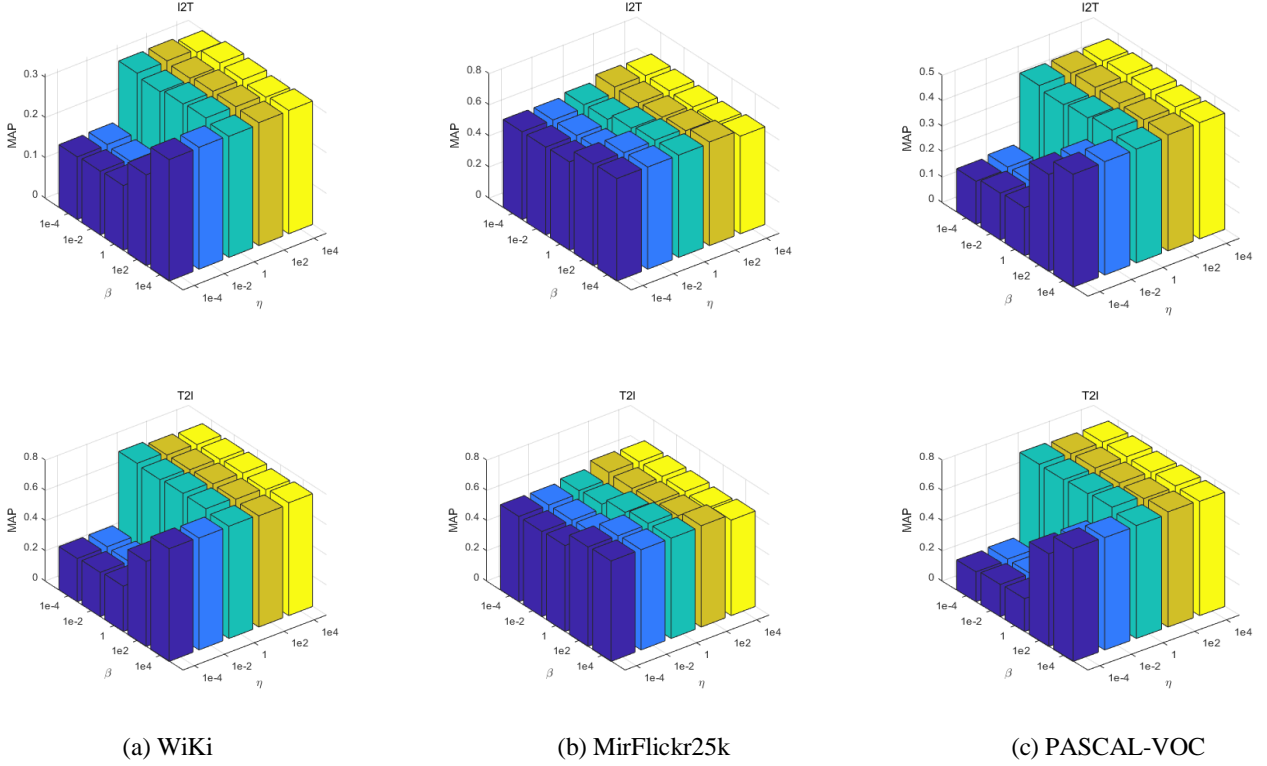

(a) WiKi  (b) MirFlickr25k  (c) PASCAL-VOC

**Fig. 3.** The MAP variations of different parameters settings on WiKi, MirFlickr25k and PASCAL-VOC. Different rows show different tasks and different columns correspond to different dataset.

*3.2. Experimental Setting*

In our experiments, we carry out cross-modal retrieval including two typical retrieval tasks. i.e., Image query the Text database and Text query the Image database which are often abbreviated to I2T and T2I respectively. We compare our method with a few state-of-art cross-modal hashing methods: SCM-orth [15], SCM-seq [15], SMFH [18], DCH-RBF [19], CCA [25], IMH [10], FSH [9]. SCM-orth, SCM-seq, SMFH, and DCH-RBF are supervised hashing learning methods which add semantic label into their framework; CCA is a classical method which maximizes the correlation between multiple modalities and IMH learns a common space for cross modal retrieval by preserving intra-media and inter-media consistence; FSH embeds an undirected asymmetric graph to learn binary codes for each modality. The Mean Average Precision (MAP) is used as the indicator of the retrieval performance. When comparing with the baselines, we empirically set $\lambda$ to be $10^{-4}$ and tune the values of $\beta$ and $\eta$ in the candidate ranges of $\{10^{-4}, 10^{-2}, 1, 100, 10^4\}$ empirically. The kernel width $\sigma$ is set according to the following rule $\sigma = 1/n^2 \sum_{i,j=1}^{n} ||x_i^{(m)} - x_j^{(m)}||^2$ for the $m$-th modality. Our experiments are implemented on MATLAB 2016b and Windows 10 (64-Bit) platform based on desktop machine with 12 GB memory and 4-core 3.6GHz CPU, and the model of the CPU is Intel(R) CORE(TM) i7-7700.

*3.3. Retrieval Performance Evaluation*

The Mean Average Precision (MAP) is adopted to evaluate the retrieval performance. A larger MAP score indicates a better performance. The Average Precision for a query $q$ is defined as follows

$$AP(q) = \frac{1}{l_q} \sum_{m=1}^{R} P_q(m) \delta_q(m) \quad (19)$$

where $l_q$ denotes the correct statistics of top $R$ retrieval results; $P_q(m)$ is the accuracy of the top $m$ retrieval results; If the result of position $m$ is correct, $\delta_q(m)$ is equal to one, and zero otherwise. In our experiments, $R$ is set to be the size of entire retrieval set.

We conduct our experiments on WiKi, MirFlickr25k and PASCAL-VOC respectively by varying the length of binary codes from 16 to 128. From Table 1 to Table 3, we can see that the performance of our method improves with the increasing length of the binary codes. Compared with the baseline methods, the proposed SDH achieves comparative results in terms of the MAP score. DCH-RBF is an effective method that does not take class-specific semantics preservation into account. The MAP results of our method exhibit average improvements of 2.6%, 2.9% and 22.9% over the DCH-RBF on WiKi, MirFlickr25k and PASCAL-VOC respectively. It is worth noting that the difference between DSH-*linear* and DSH is that DSH-*linear* linearly maps different modalities into the common space. As shown in Table 1, Table2 and Table3, DSH is superior to DSH-*linear*. Especially on the task T2I, the MAP results of DSH are better than DSH-*linear* by approximately 48% and 37% on WiKi and PASCAL-VOC respectively.

3.4. Retrieval Parameter Analysis

In our framework, $\beta$ and $\eta$ are two adjustable parameters which control the weight of the preservation of discriminative information and of class-specific semantic content respectively. Fig. 3 shows sensitivity of the result to $\beta$ and $\eta$ when using 128-bit hash code on WiKi, MirFlickr and PASCAL-VOC. We can observe that the variation of each of the two parameters can influence the retrieval performance when fixing the other. Thus it can be seen that the preservation of discriminative information and class-specific semantics work together to boost the retrieval performance. Further, we conduct experiments to analyze how the sample size $M$ affects the retrieval performance. Table 4 shows I2T, T2I and training time on the WiKi, MirFlickr25k and PASCAL-VOC when increasing the sampling size M from 100 to 2,100 and the length of hashing code is fixed to 16 bits. From Table 4, we can see that the performance tends to be improved with the increasing sample size. However, the training and predication cost rise. Thus, the choice of sample size should balance the retrieval performance and computational costs.

*3.5. Convergence Analysis and Computational Complexity*

Each variable of Algorithm 1 is updated iteratively until convergence. The convergence curves on the WiKi, MirFlickr25k and PASCAL-VOC are plotted in Fig. 2. As shown in Fig. 2, our method converges quickly, although it is difficult to prove the theoretical convergence of the proposed algorithm.

The time complexity of DSH consists of two parts: computing kernelized matrix and the training phase. The time complexity of computing kernelized matrix is $O(nMd)$ where $d = \max(d_m | m = 1,2 \dots)$. The complexity of each iteration is $O(M^2 n + tncr^2)$ where $t$ is the iteration number of DCC. Thus, the overall complexity of DSH is $O((M^2 + tcr^2)nT + nMd)$ where $T$ is the iteration number of the algorithm 1.

## 4. Conclusion

In this paper, we propose a new model (DSH) which integrates subspace learning, classifier learning and the basis matrix learning into a joint framework to learn the unified hashing features that both retain discrimination ability and preserve class-specific content by using the label matrix. In contrast to previous works, a non-linear method is introduced to learn a common subspace. We adopt the efficient DCC algorithm to optimize the problem with discrete constraint. We evaluate our method on three benchmark datasets and the results show the effectiveness of our method. In the future, we plan to study the theoretical convergence of the proposed algorithm and adopt DNN features.

## Acknowledgments

THE PAPER IS SUPPORTED BY THE NATIONAL NATURAL SCIENCE FOUNDATION OF CHINA(GRANT NO.61373055, 61672265), UK EPSRC GRANT EP/N007743/1, MURI/EPSRC/DSTL GRANT EP/R018456/1, AND THE 111 PROJECT OF MINISTRY OF EDUCATION OF CHINA (GRANT NO. B12018)